\documentclass{article}

\usepackage{microtype}
\usepackage{graphicx}
\usepackage{subfigure}
\usepackage{booktabs} 

\usepackage{hyperref}

\usepackage{hyperref}       
\usepackage{url}            
\usepackage{booktabs}       
\usepackage{amsfonts}       
\usepackage{nicefrac}       
\usepackage{microtype}      
\usepackage{graphicx}
\usepackage{cite}
\usepackage{mathtools}

\usepackage{verbatim}
\usepackage{listings}
\usepackage{multirow}      
\usepackage{multicol} 
\usepackage{wrapfig}

\newtheorem{definition}{Definition}

\usepackage{algorithmic}

\newcommand{\ignore}[1]{}

\usepackage{listings}
\usepackage{xcolor}

\usepackage{hyperref}

\def\BibTeX{{\rm B\kern-.05em{\sc i\kern-.025em b}\kern-.08em
    T\kern-.1667em\lower.7ex\hbox{E}\kern-.125emX}}

\usepackage[ruled,vlined]{algorithm2e}
 
\usepackage{algorithmic}
\usepackage{etoolbox}
\newcommand{\algorithmicdoinparallel}{\textbf{do in parallel}}
\makeatletter
\AtBeginEnvironment{algorithmic}{%
  \newcommand{\FORALLP}[2][default]{\ALC@it\algorithmicforall\ #2\ %
    \algorithmicdoinparallel\ALC@com{#1}\begin{ALC@for}}%
}
\makeatother

\usepackage{enumitem}

\setlist[enumerate]{itemsep=-1mm}

\usepackage{amsmath}

\usepackage[accepted]{icml2020}

\begin{document}

\twocolumn[
\icmltitle{PrivacyFL: A simulator for privacy-preserving and secure federated learning}



\icmlsetsymbol{equal}{*}

\begin{icmlauthorlist}
\icmlauthor{Vaikkunth Mugunthan}{mit}
\icmlauthor{Anton Peraire-Bueno}{mit}
\icmlauthor{Lalana Kagal}{mit}

\end{icmlauthorlist}

\icmlaffiliation{mit}{CSAIL, Massachusetts Institute of Technology, Cambridge, MA, USA}

\icmlcorrespondingauthor{Vaikkunth Mugunthan}{vaik@mit.edu}
\icmlcorrespondingauthor{Anton Peraire-Bueno}{aperaire@mit.edu}
\icmlcorrespondingauthor{Lalana Kagal}{lkagal@mit.edu}

\icmlkeywords{Machine Learning, ICML}

\vskip 0.3in
]



\printAffiliationsAndNotice{}  

\begin{abstract}
Federated learning is a technique that enables distributed clients to collaboratively learn a shared machine learning model without sharing their training data. This reduces data privacy risks, however, privacy concerns still exist since it is possible to leak information about the training dataset from the trained model’s weights or parameters. Therefore, it is important to develop federated learning algorithms that train highly accurate models in a privacy-preserving manner. Setting up a federated learning environment, especially with security and privacy guarantees, is a time-consuming process with numerous configurations and parameters that can be manipulated. In order to help clients ensure that collaboration is feasible and to check that it improves their model accuracy, a real-world simulator for privacy-preserving and secure federated learning is required. 

In this paper, we introduce PrivacyFL, which is an extensible, easily configurable and scalable simulator for federated learning environments. Its key features include latency simulation, robustness to client departure/failure, support for both centralized (with one or more servers) and decentralized (serverless) learning, and configurable privacy and security mechanisms based on differential privacy and secure multiparty computation (MPC). 

In this paper, we motivate our research, describe the architecture of the simulator and associated protocols, and discuss its evaluation in numerous scenarios that highlight its wide range of functionality and its advantages. Our paper addresses a significant real-world problem: checking the feasibility of participating in a federated learning environment under a variety of circumstances. It also has a strong practical impact because organizations such as hospitals, banks, and research institutes, which have large amounts of sensitive data and would like to collaborate, would greatly benefit from having a system that enables them to do so in a privacy-preserving and secure manner. 
\end{abstract}

\section{Introduction}
Federated learning has become an important area of research as it provides an optimal solution to train a collaborative model when data is dispersed across different organizations. Its primary advantage is that it allows statistical models to be trained over remote entities while keeping data localized. For example, individual hospitals can collaboratively train a machine learning model in order to improve their predictive accuracy for a given task. However, most clients are unwilling to share their data or inadvertently leak information due to internal privacy policies or regulation. Therefore, it is important to develop federated learning frameworks that train highly accurate models in a privacy-preserving manner.



\begin{figure}
    \centering
    \includegraphics[width=8cm]{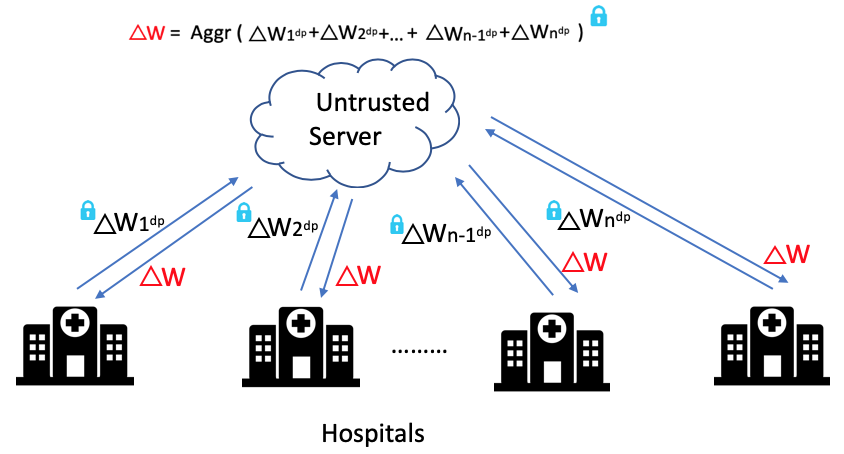}
    \caption{Secure and Privacy-Preserving Federated Learning}
    \label{fig:DP}
\end{figure}

Let us assume that there are several hospitals that want to develop a shared machine learning model. First, each hospital trains a local model, $\Delta W_i$, rather than sending raw data to a centralized server. Hospitals act as remote clients and communicate with a central server at regular intervals to learn a global model. For every iteration, the hospitals send their local model to the server. The server computes and sends back the global model to individual hospitals. This process repeats until convergence occurs or some sort of stopping criterion is achieved. There are two main drawbacks in this approach: (i) communication is not secure, and (ii) collusion of $n-1$ clients in a $n$ client system reveals the model of the remaining client. The use of MPC helps with the first issue and the second one requires a privacy-preserving mechanism such as differential privacy. This is illustrated in Figure \ref{fig:DP}. The differentially private inputs, $\Delta W_i^{dp}$, are encrypted and sent to the server. The server aggregates the inputs received from the clients and sends them back to the client.


Some of the core challenges of federated learning include:
\begin{itemize}
    \item \textbf{Communication cost:} This is a huge bottleneck as it increases as the number of iterations increases and becomes an issue as the number of clients increases. Using the simulation results, a system designer can decide whether it is feasible to participate in the federated learning environment. 
    
    
    \item \textbf{Privacy and security guarantees:} In a federated learning environment, a malicious client can compromise the privacy of individuals in the training data by inferring details of a training dataset from the model's weights  ~\cite{shokri2017membership,nasr2018comprehensive}. These privacy issues can be addressed through the use of differential privacy and MPC. Our simulator allows the system designer to evaluate the performance with differential privacy and security mechanisms. 
\end{itemize}

While designing a federated learning environment, it is important to understand the trade-offs between privacy, efficiency, and accuracy. System designers should be able to choose different parameters and compare multiple scenarios in order to find the right trade-offs for their environment. Simulators are essential for federated learning environments for the following reasons:
\begin{itemize}
    \item To evaluate accuracy: It should be possible to simulate federated models and compare their accuracy with local models.
    \item To evaluate total time taken: Communication between distant clients can become expensive. Simulations are useful for evaluating if client-client and client-server communications are beneficial.  
    \item To evaluate approximate bounds on convergence and time take for convergence.
    \item To simulate real-time dropouts: Clients in a federated environment may drop out at any time. 
\end{itemize}

%


Our simulator, PrivacyFL, is designed to specifically address these issues. It is efficient in terms of speed and execution time as client and server operations happen in a parallelized manner. It provides numerous differentially private mechanisms including the Laplacian mechanism \cite{dwork2006calibrating}, the Gaussian mechanism \cite{dwork2006calibrating}, and the Staircase mechanism \cite{geng2015staircase}. System designers can choose from any of these mechanisms to ensure that the learning is privacy-preserving. Our system also provides the option for clients to use a Diffie-Hellman key exchange for inbuilt security. The simulator can also easily support the addition of new privacy-preserving algorithms. It is composed of generalized classes and functions that make it easy for developers to add/modify new algorithms and protocols. Along with simulating federated environments that use one or more central servers, our simulator is also able to  simulate a completely decentralized environment, i.e., a federated learning system without a centralized server. In the real world, environments are dynamic with clients failing, dropping off, or being added at run-time. Being able to model this is an important requirement of a simulator for federated learning and PrivacyFL is able to handle such dynamic cases. It is able to simulate computation and communication times to evaluate how long a client would have to wait for each step of the protocol to complete in a real world setting. 









The remainder of this paper is organized as follows. In Section 2, we discuss existing frameworks in the field of privacy-preserving secure federated learning. Section 3 describes background work and foundational concepts. We provide a detailed explanation of our architecture, classes, and life-cycle in Section 4. In Section 5, we discuss privacy-preserving and secure aggregation algorithms. Experiments and evaluations under different scenarios are described in Section 6, and Section 7 discusses the reproducibility of our experiments. We conclude with a summary and a discussion of our future work in Section 8.

\section{Related Work}
Federated learning is an approach to train machine learning models on distributed data. The federated averaging algorithm proposed by \cite{mcmahan2017federated} follows a synchronous training approach in which each client trains a local model and sends it to a centralized server, which computes the average of all models. The server sends the average model back to the clients. Numerous approaches have been proposed to enhance privacy and security in a federated learning environment. \cite{mcmahan2017learning} applied differential privacy to provide the necessary privacy guarantees whereas \cite{bonawitz2017practical} proposed a secure aggregation protocol for privacy-preserving machine learning in a federated environment.


\cite{bonawitz2019towards} built a production system, the first of its kind, for federated learning in the field of mobile devices using TensorFlow. The authors use the Federated Averaging algorithm proposed by \cite{mcmahan2017federated}. However, their system has the following limitations, (i) though it does support privacy, it does not allow for new mechanisms to be added easily,  (ii) it is dependent on TensorFlow so you cannot use it with other machine learning libraries, (iii) it does not focus on simulating virtual environments with multiple servers and multiple clients at different locations, and (iv) it is not very efficient because it does not support the parallel execution of training sessions.

Our simulator, PrivacyFL, was specifically developed to overcome these issues by providing a highly configurable system. For more details, please see Section 4. 

There are a myriad of challenges pertaining to privacy in federated learning. In addition to guaranteeing privacy, it is important to make sure that the communication cost is cheap and efficient. There are numerous privacy definitions for federated learning \cite{geyer2017differentially}\cite{bhowmick2018protection}  \cite{thakkar2019differentially}. We can classify them into 2 categories: local privacy and global privacy. In local privacy, each client sends a differentially private value that is secure/encrypted to the server. In the global model, the server adds differentially private noise to the final output.

MPC and differential privacy  are  the most common techniques to guarantee security and privacy in a federated environment setup \cite{bhowmick2018protection} \cite{mcmahan2017learning}\cite{bonawitz2017practical}. 


In a federated learning setup, instead of sharing data, clients share models. Though this seems to provide increased privacy, there have been a multitude of privacy attacks, including reconstruction and membership inference attacks \cite{shokri2017membership}\cite{nasr2019comprehensive}\cite{melis2019exploiting}, demonstrating that additional privacy is required in the form of protecting model parameters. Cryptographic techniques such as secure multiparty computation (MPC) \cite{Yao86,CCD87,goldreich1998secure} guarantee that clients don't learn anything except the final cumulative model weight.  Also, these techniques do not compromise the accuracy of models. MPC allows clients to collaborate in order to compute a function without clients revealing their individual (and possibly sensitive) inputs to other clients.

Though MPC reveals only the final result and not the individual inputs of individual clients, sometimes, this information is sufficient to infer the individual input of a client. For example, let us consider the scenario where $n-1$ clients collude against an honest client in a $n$ client federated environment and the clients want to compute the overall sum.
Once the sum is known, the clients can identify the private input of the non-colluding client.  In order to overcome this, differential privacy \cite{dwork2006calibrating} can be used in addition to MPC. This solution is resistant to collusion attacks and Figure \ref{fig:DP} represents a privacy-preserving secure federated learning environment.

The above-mentioned methods are not dependent on each other and can exist independently in a federated learning environment. Our simulator allows the system designer to choose and configure different privacy and security mechanisms according to their needs and provides a flexible architecture.

Our system provides necessary guarantees for the "honest-but-curious" threat model. In this threat model, all the clients in the system follow the protocol but try their best to identify the data/model of other clients. As part of our future work, we plan to add solutions to the "malicious" threat model where clients need not follow the protocol and may conduct adversarial attacks.

\section{Background}
\subsection{Differential Privacy}
PrivacyFL provides several differential privacy mechanisms that can be used and configured by the system designer. Differential privacy \cite{dwork2006calibrating} is a mathematical privacy framework that captures the notion of risk when an individual is included in a dataset. A detailed explanation of differential privacy and the associated mechanisms used in the paper is provided in Appendix \ref{diffpriv}.

\subsection{Differentially Private Federated Logistic Regression }
\label{fedLogReg}
We use differentially private logistic regression for our experiments. A detailed explanation of differentially private federated logistic regression is provided in Appendix \ref{fedLogReg}.

\section{Architecture }

The purpose of our simulator is to provide a light-weight but efficient Python framework that enables clients to simulate privacy-preserving secure federated learning. In a simulation, a client corresponds to an instance of the \textbf{ClientAgent} class. Each client agent is capable of interacting with the other client agents to establish common keys before the simulation. Client agents also communicate with an instance of the \textbf{ServerAgent} class if configured to be a centralized system. The server agent is an agent who is not training its models but instead serves to run the federated learning algorithm. It is responsible for requesting weights from the clients, averaging the weights, and returning to the clients the federated weights. Many applications will only require one server agent, but it is worth noting that with some modifications the framework can handle any number of server agents. 

To run the simulation, an instance of the \textbf{Initializer} class must be created. The initializer will create the agents in the simulation and also invoke the client agents to perform any offline-stage logic such as key exchanges. The initializer also gives each agent a directory, which contains a mapping of agent names to agent instances so that agents can invoke each others methods, thereby enabling all agents to communicate. The initializer also gives each client agent an instance of the \textbf{ModelEvaluator} class, which it uses to evaluate the federated learning on a test dataset. 


\begin{figure}
    \centering
    \includegraphics[width=0.5\textwidth]{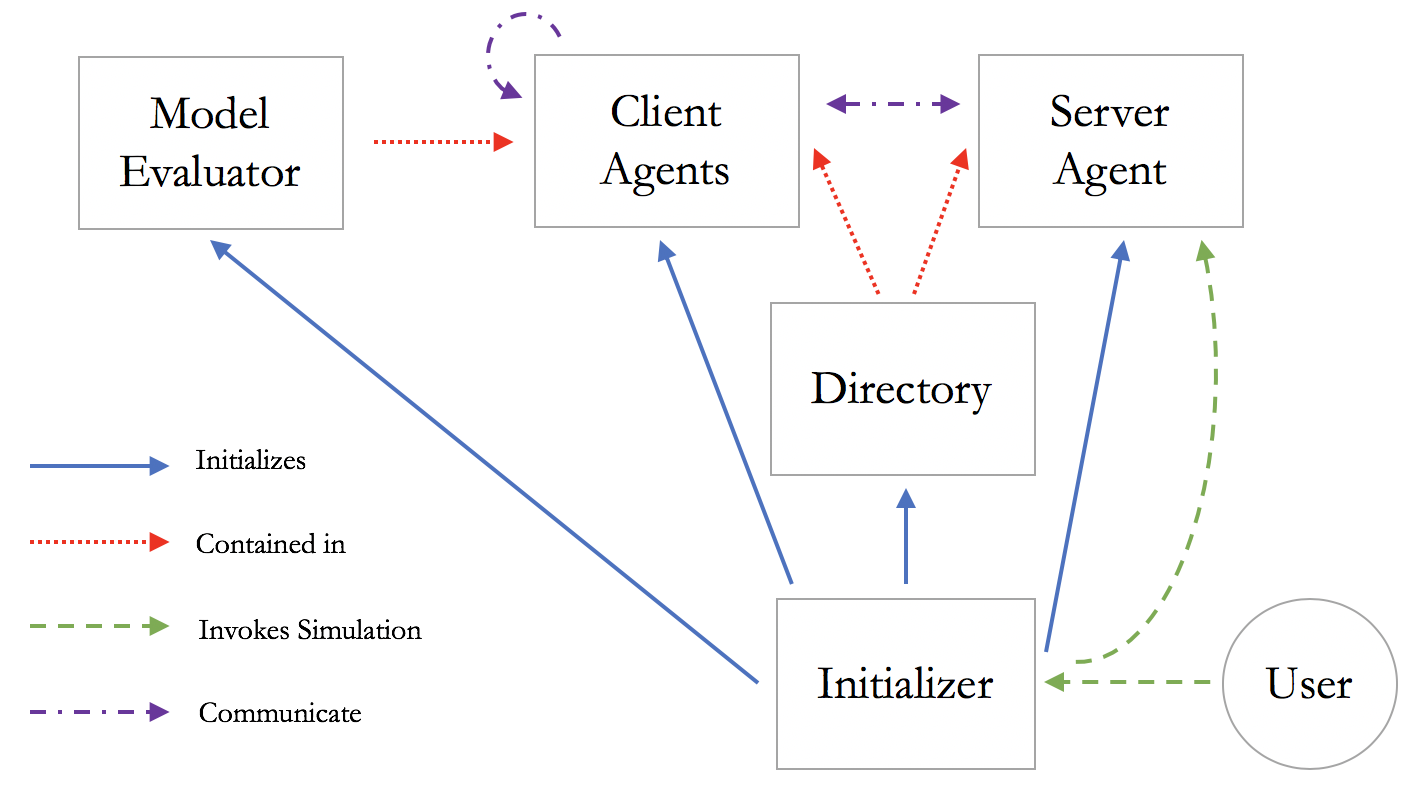}
    \caption{System diagram containing relationships between the important classes in PrivacyFL}
    \label{SystemArchitecture}
\end{figure}

\subsection{Simulation Lifecycle}
This section provides an overview of the steps carried out in a simulation. 

\begin{itemize}
    \item [1.] System designer wishing to simulate a federated learning environment specifies simulation parameters in \path{config.py} and runs \path{run_simulation.py}.
    \item [2.] \path{run_simulation.py} creates an instance of Initializer, which instantiates the client agents and server agent. It also creates a mapping of agent names to instances and passes those to each client.
    \item [3.] \path{run_simulation.py} invokes the initializer's \path{run_simulation()}, which subsequently calls the server agent's \\ \path{request_values()} method.  
    \item [4.] Repeat for $i = 1, 2, 3, ... \lstinline{num_iterations}$:
        \begin{itemize}
            \item [4.1] Server agent requests weights from clients in parallel by invoking each client's \path{produce_weights()} method.
            \item [4.2] In \path{produce_weights()}, each client trains its machine learning model on $D_i$, the dataset for that iteration. Each client saves the weights locally, and then creates a copy of its weights to which it adds a security offset and differentially private noise. It returns the modified weights to the server. 
            \item [4.3] Upon receiving the weights from each client, the server averages the weights and returns the federated weights to each client in parallel through their \path{receive_weights()}. 
            \item [4.4] Upon receiving the federated weights for iteration $i$ from the server, each client computes the accuracy of the federated model vs its local model on the test-set. Each client then computes whether its weights have converged and tells the server whether it is dropping out after this iteration.
            \item [4.5] To conclude the iteration, the server tells all the clients which other clients have dropped out of the simulation.
        \end{itemize}
\end{itemize}

\subsection{Classes}
There are 6 main classes in the simulator namely \textbf{Message}, \textbf{Agent}, \textbf{ClientAgent}, \textbf{ServerAgent}, \textbf{Initializer}, and \textbf{Directory}.

\subsubsection{The Message Class}
All agent-agent communications occur by one agent invoking another agent's method with a message as the sole argument. The message class contains metadata about the communication as well as a \path{body} attribute, which is a dictionary containing all the values an agent is sending. 

\subsubsection{The Agent Class} 
The Agent class is not meant to be initialized directly. Rather, its intended usage is to be sub-classed to create agents with more specific behavior. It is the base class for the ClientAgent and ServerAgent classes we provide, but can be sub-classed to create a different kind of agent if ClientAgent and ServerAgent are not easily modified to fit the needs of your simulation. 

\subsubsection{The ClientAgent Class} 
An instance of the ClientAgent class is an entity that is training a machine learning model on the same task as the other client agents. Client agents are assigned an \path{agent_number}, which is then appended to the string \path{client_agent} to create their name. For example, if there are three clients they are named client\_agent0, client\_agent1, and client\_agent2. There are two important public methods of ClientAgent that are invoked by the ServerAgent in the online portion of the simulation. 

\begin{itemize}
\item \path{produce_weights(self, message)} is called every iteration and prompts the client to train its machine learning model on its dataset for that iteration. The message contains the iteration, a mutex lock used for multi-threading, and information about the simulated time.
\item \path{receive_weights(self, message)} is called every iteration when the server has federated weights to return to the client. The message contains the iteration, weights, and the simulated time this message is received. This method returns \path{True} if the client agent's weights have converged with the federated weights and \path{False} otherwise.
\item \path{remove_active_clients(self, message)} is called at the end of the iteration. The message contains the iteration, list of clients that have dropped out, and the simulated time the message is received. This method is only used if the simulation is configured with client drop out.
\end{itemize}

In addition, client agents have some methods for the  client-client communication that are used to establish shared common keys with the other clients. 
\begin{itemize}

\item The \path{send_pubkeys(self)} of each client is invoked by the initializer once all client agents have been initialized. In this method, each client computes secrets $a_{i,1},\ldots,a_{i,n}$ and sends $a_{i, j}$ to client $j$ by invoking that client's \path{receive_pubkey()} method.
\item In \path{receive_pubkey(message)}, each client saves the public key sent to it by the other clients. 
\item \path{initialize_common_keys(self)} is called once all clients have exchanged public keys using the above two methods. In this method, the common key list is initialized. This common key list will be used to produce security offsets for the online portion of the simulation. 
\end{itemize}

\subsubsection{The ServerAgent Class} 
An instance of the ServerAgent class represents a third-party entity that coordinates the online portion of the simulation. ServerAgent has one method: \begin{itemize}
    \item \path{request_values(num_iterations)} is called by an instance of Initializer to signal the server agent to start requesting values from clients.  \path{request_values()} first requests weights in parallel from the clients by calling their \path{produce_weights()}. It then averages these weights and returns them to the clients in parallel by calling their \path{receive_weights()} method. Finally, if the simulation is configured to allow client dropout, the server invokes the clients' \path{remove_active_clients()} to indicate which clients have dropped out at the end of this iteration. 
\end{itemize}

\subsubsection{The Initializer Class}
An instance of the Initializer class is used to initialize the agents and model evaluator. In addition any offline stage logic, such as prompting the clients to perform a Diffie-Hellman key exchange, should occur in this class. In our example, it loads the MNIST dataset \cite{lecun1998mnist}, partitions it, and distributes the correct amount of data to each client.
To commence the simulation, one creates an instance of the Initializer class and invokes its \path{run_simulation()} method, which then invokes the server agent's \path{request_values()} to commence the online portion of the simulation.

\subsubsection{The Directory Class} 
An instance of the Directory class contains a mapping of agent names to agent instances that allows agents to invoke other agents' methods by only having their name. An instance of Directory is created in the \path{__init__} method of the Initializer class after all the agents have been created. It is then passed on to all the agents using their \path{set_directory()} method.


\subsection{Configurations and Features}
\label{Config}
PrivacyFL can be configured to simulate a variety of scenarios with the configuration parameters available in \path{config.py}. These configuration parameters are described in Table \ref{table:config} in Appendix \ref{configParams}.

In addition, config.py allows system designers to set several other parameters such as the number of clients, custom $\epsilon$ and dataset sizes for each client, and thread-safe random seed for reproducibility.

\section{Use case}
To evaluate our simulator, we developed a secure and differentially private federated logistic regression algorithm. 
\subsection{Differentially Private Federated Averaging}
The federated averaging algorithm is one of the most popular approaches of federated learning. It is a customized version of parallel Stochastic Gradient Descent. In our approach, Algorithm \ref{alg:GoogleclientDP}, each client runs $U$ rounds of local SGD and returns the trained weights. Clients make use of new data every round. Here, each client adds the difference of gamma random variables to their unperturbed weights. In the aggregation phase, the server adds the perturbed weights it receives from all values. The server aggregates the weights into the final model, which are differentially private because
    \begin{center}
    $L(\mu,\lambda) = \frac{\mu}{n} + \sum_{k=1}^{n} \gamma_k - \gamma'_k$
\end{center}

$\gamma_k$ and $\gamma'_k$ are Gamma distributed random variables. 

The client-server communication happens $R$ times unless a client drops out.

We propose and make use of two algorithms for evaluation and their explanations are provided in Appendix \ref{algorithms}. We propose Algorithm \ref{alg:cumulativeDP}, where clients reuse old data used in previous rounds and also make use of any new data available. 

The main difference between Algorithm \ref{alg:GoogleclientDP} and Algorithm \ref{alg:cumulativeDP} is that, in Algorithm \ref{alg:cumulativeDP} clients retrain their model from scratch on their entire local dataset, which may or may not get updated every round, and don't use the weights received from the server while running local gradient descent. In Algorithm \ref{alg:GoogleclientDP}, clients compare their weights from the previous iteration with that of the global weights they receive from the server. With all algorithms, if local convergence is achieved then the client can drop out of the system.

In all the algorithms, when clients receive the federated weights from the server they are able to subtract the differentially private noise that they contributed to the federated weights.

\subsection{Secure Aggregation}
To provide the necessary security guarantees, we use a combination Diffie-Hellman key exchange and Pseudo-Random Generators to ensure a secure aggregation protocol. Our protocols ensure that clients only need to communicate with each other once at the start of the simulation to ensure security for all iterations, thereby reducing the amount of client-client communications and overall communication time.

\section{Experiments}
\label{exp}
For our evaluation, we configured the clients to train on UCI's Machine Learning MNIST dataset \cite{lecun1998mnist} \cite{newman1998uci} for eight iterations. This dataset is easily loaded using scikit-learn's \path{datasets.load_digits()} module. We ran the experiments on a MacBook Pro 2.3 GHz Intel Core i5, using Python's multithreading module for parallelization. All experiments were initialized with the same thread-safe random seeds, ensuring fair comparisons where possible. The experiments and their evaluation can be found in Appendix \ref{experiments}.

\section{Availability}

\begin{figure}[h!]
    \centering
    \includegraphics[width=\linewidth]{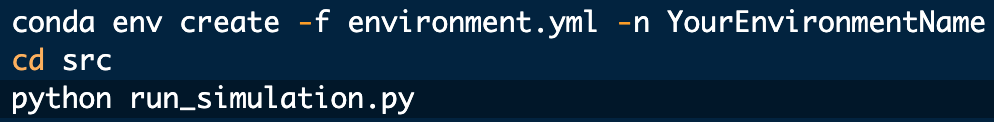}    
    \caption{Steps to execute the code}
    \label{howtorun}
\end{figure}

PrivacyFL is available on GitHub at \url{https://github.com/vaikkunth/PrivacyFL/} under the MIT License. To run the simulator, the repository should be cloned locally and a conda environment with the necessary dependencies should be created as shown in Figure \ref{howtorun}. The config.py file contains a variety of parameters as described in Section \ref{Config} that allow the simulation to be customized. \path{utils/data_formatting.py} provides an an example of how to pre-process dataset for the simulationusing the MNIST dataset as an example. In addition, the \path{config.py} file also has random seeds for each client that ensure that randomness is handled in a reproducible thread-safe manner.

\vspace{-0.2cm}
\section{Conclusion and Future Work}

In this paper, we presented PrivacyFL, a simulator for privacy-preserving and secure federated learning. The primary features of our system include latency simulation, robustness to client departure/failure, support for both server-based and serverless federated learning, and tunable parameters for differential privacy and MPC.

We evaluated our simulator by running multiple experiments on federated logistic regression to demonstrate the flexibility of our framework. Our simulator and protocols can be easily applied to other federated machine learning algorithms as well.  PrivacyFL is highly customizable and system designers can easily configure parameters such as setting different latencies, choosing appropriate differential privacy mechanisms, and simulating clients dropping out or being added dynamically.

As part of our future work, we plan to explore solutions to the "malicious" threat model where clients need not follow the protocol and may conduct adversarial attacks. 
We plan to extend our simulator to real-time cluster environments and perform experiments on larger datasets. Once our simulator is capable of being deployed in a cloud computing platform, we plan to work with pilot partners for the real-time testing of our system and protocols.

\bibliography{example_paper}
\bibliographystyle{icml2020}

\appendix
\section{Differential Privacy}
\label{diffpriv}

\begin{definition}
    ($(\epsilon, \delta)$-Differential Privacy) A randomized mechanism $\mathcal{M}$ satisfies $(\epsilon,\delta)$-differential privacy ($(\epsilon,\delta)$-DP) when there exists $\epsilon > 0$, $\delta > 0$, such that 
    $$\text{Pr [}\mathcal{M}(D_1)\in S\text{]}\leq e^\epsilon \text{Pr [}\mathcal{M}(D_2)\in S\text{]} + \delta$$
    holds for every $S \subseteq $ Range($\mathcal{M}$) and for all datasets $D_1$ and $D_2$ differing on at most one element. 
\end{definition}

When $\delta=0$, we say that the randomized mechanism $\mathcal{M}$ satisfies $(\epsilon,0)$-differential privacy

\begin{definition}
    (Global Sensitivity) For any real-valued query function $t: \mathcal{D} \rightarrow \mathbb{R}$, where $\mathcal{D}$ denotes the set of all possible datasets, the global sensitivity $\Delta$ of $t$ is defined as $$\Delta = \max_{\mathcal{D}_1\sim\mathcal{D}_2} |t(\mathcal{D}_1)-t(\mathcal{D}_2)|,$$ for all $\mathcal{D}_1\in\mathcal{D}$ and $\mathcal{D}_2\in\mathcal{D}$.
\end{definition}

\subsubsection{Laplacian Mechanism}

The Laplacian mechanism preserves $\epsilon$-differential privacy \cite{dwork2006calibrating}. We make use of the random noise $X$ drawn from the symmetric Laplacian distribution. The zero-mean Laplacian distribution has a  probability density function $f(x)$.
$$f(x) = \frac{1}{2\lambda}e^{-\frac{|x|}{\lambda}},$$ where $\lambda$ is the scale parameter.
Given $\Delta$ of the query function $t$, and the privacy loss parameter $\epsilon$, the \textit{Laplacian mechanism} $\mathcal{M}$ uses random noise $X$ drawn from the Laplacian distribution with scale $\lambda = \frac{\Delta}{\epsilon}$.

\subsubsection{Distributing Differentially Private Noise}
	In addition to the basic differential privacy mechanisms, where each client adds differentially private noise, our framework also supports distributed differentially private noise addition. In distributed differential privacy, each client contributes a portion of the total differentially private noise. The sum of all these individual contributions result in a differentially private noise value. 
	
	As the Laplace mechanism satisfies differential privacy, we show how a Laplace Random variable can be generated from Gamma random variables. 
	
	\begin{center}
    $L(\mu,\lambda) = \frac{\mu}{n} + \sum_{p=1}^{n} \gamma_p - \gamma'_p$,
\end{center}
where $\mu$ and $\lambda$ are the mean and scale parameters of the Laplacian mechanism.

$\gamma_p$ and $\gamma'_p$ are Gamma distributed random variables with probability density functions defined as :
\begin{center}
    $\frac{(1/s)^{1/n}}{\Gamma(1/n)}x^{1/n -1}e^{-x/s}$,
\end{center}
where $1/n$ is the shape parameter, $s$ is the scale parameter and $\Gamma(p)= \int_{0}^{\infty} x^{p-1}e^{-x}dx $

We make use of this technique where each client adds $\gamma_p - \gamma_p'$ in our distributed Algorithm \ref{alg:GoogleclientDP}.

\section{Differentially Private Federated Logistic Regression }
\label{fedLogReg}
We use differentially private logistic regression for our experiments.

In a local differentially private output perturbation setting, each client adds differentially private noise to their model parameters. In this setting, each  client adds Laplacian noise to their model weights and participates in federated learning. Once clients train private models, they coordinate with the server to compute the global model. This client-server communication is repeated till convergence or a predefined number of rounds.

Let  $\hat{w_i}^{dp}$, for $i \in 1$ to $n$, depict the local model constructed by each client after minimizing the objective function and adding differentially private noise to their weights. Here $w$ represents the vector of weights and $n$ represents the number of clients.

$\hat{W}=\frac{1}{n}\sum_{i=1}^{n}\hat{w_{i}}^{dp}$ is the differntially-private cumulative model.

In a trusted server setting, every client computes its model and sends it to the server. The server aggregates the model inputs it receives from clients and adds differentially private noise to the average. Here,

$\hat{W}=\frac{1}{n}\sum_{i=1}^{n}\hat{w_i} + \eta$, $\eta$ is the differentially private noise added to the cumulative model.

We also consider the case where each client contributes to a portion of the differentially private noise in the global setting. That is each client adds  $\hat{w_i}^{dpPartial}$ to its model parameters, for $i \in 1$ to $n$. The server does the following, $\hat{W}=\frac{1}{n}\sum_{i=1}^{n}\hat{w_{i}}^{dpPartial}$.

For logistic regression in the federated multi-client setting, \cite{jayaraman2018distributed} proposed the following:
\mathchardef\mhyphen="2D
 For $1$-$Lipschitz$ , $\Delta=\frac{2}{n*k*\alpha}$ for a federated multi-client setting. Here $k$ is the size of the smallest dataset amongst the $n$ clients and $\alpha$ is the regularization parameter. Hence, $\eta=Laplace(\frac{2}{n*k*\alpha*\epsilon})$, where $\epsilon$ is the privacy loss parameter.

\section{Algorithms}
\label{algorithms}

\begin{algorithm}
\SetAlgoLined
  \textbf{Server Implements:\\}
  \begin{algorithmic}[]
    \STATE  Initialize $w_0$
    \FOR{$r \gets 1$ to $R$}
      \FOR{every client $k \in n$ \textbf{in parallel}}
        \STATE $w^{k}_{r+1} \gets ClientProcedure(k,w_{r})$
      \ENDFOR
      \STATE $w_{r+1} \gets \frac{1}{n}\sum_{i=1}^{n} w^{i}_{r+1} $
    \ENDFOR
  \end{algorithmic}
  \textbf{ClientProcedure(k,w): \\}
  \begin{algorithmic}[]
    \FOR{local gradient update  $u \gets 1$ to $U$}
        \STATE $w \gets w - \eta \nabla g(w)$
    \ENDFOR\\

    Return $w+\gamma -\gamma'$

  \end{algorithmic}
  
   \caption{Differentially Private Federated Averaging - Each client contributes a portion of the total differentially private noise}
   \label{alg:GoogleclientDP}
\end{algorithm}

\begin{algorithm}
\SetAlgoLined
  \textbf{Server Implements:\\}
  \begin{algorithmic}[]
    \FOR{$r \gets 1$ to $R$}
      \FOR{every client $k \in n$ \textbf{in parallel}}
        \STATE $w^{k}_{r+1} \gets ClientProcedure(k,w_{r})$
      \ENDFOR
      \STATE $w_{r+1} \gets \frac{1}{n}\sum_{i=1}^{n} w^{i}_{r+1} $
    \ENDFOR
  \end{algorithmic}

  \textbf{ClientProcedure($k,w_s$): \\}
  
  \begin{algorithmic}[]
  \IF{Round 1}\STATE Initialize $w_c^{dp}=0$
  \ENDIF
    \IF{$w_s < w_c^{dp}$}
        \STATE  Initialize $w$
        \FOR{local gradient update  $u \gets 1$ to $U$}
        \STATE $w \gets w - \eta \nabla g(w)$
    \ENDFOR\\
    $w_c^{dp} \gets w+\gamma -\gamma'$
    
    \ENDIF\\
    Return $w_c^{dp}$

  \end{algorithmic}

   \caption{Differentially Private Weighted Averaging }
     \label{alg:cumulativeDP}
\end{algorithm}

\section{Experiments}
\label{experiments}

\subsection{Experiment 1:  Accuracy vs Privacy vs Number of Clients Trade-offs}
\begin{figure}[h!]
    \centering
    \includegraphics[width=8cm]{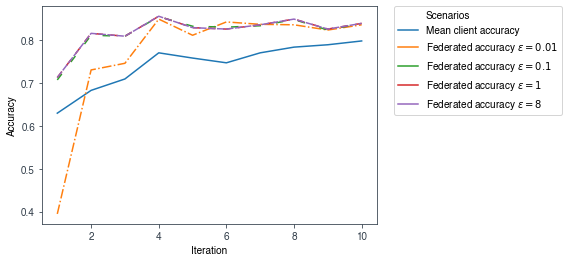}    
    \caption{Mean client accuracy on test dataset versus federated model accuracy for different $\epsilon$. Clients follow Algorithm \ref{alg:cumulativeDP}.}
    \label{Results1}
\end{figure}

One of the principal uses of the simulator is to enable clients to assess whether the accuracy of their model would increase if they participated in the federated learning. For that reason, we compare the mean accuracy of the clients' model with the accuracy of the federated global model for each iteration. We used Algorithm \ref{alg:cumulativeDP}. When computing a client's accuracy on iteration $i$, we consider its weights with no differential privacy since this scenario corresponds to clients not participating in the federated learning, thus having no need to add differentially private noise. Figure \ref{Results1} compares the mean of three clients' accuracy with the federated accuracy for different values of $\epsilon$, where a smaller value of $\epsilon$ corresponds to less noise. As expected, the federated model's accuracy increases as $\epsilon$ increases, but there is not a significant difference between $\epsilon = 1$ and $\epsilon = 8$. It is also important to note that the federated accuracy for the $\epsilon=0.1$ line increases more rapidly than the mean client accuracy. This is expected behavior since the amount of noise added per client is smaller as the size of each client's dataset increases by thirty samples each iteration.

\begin{figure}[h!]
    \centering
    \includegraphics[width=8cm]{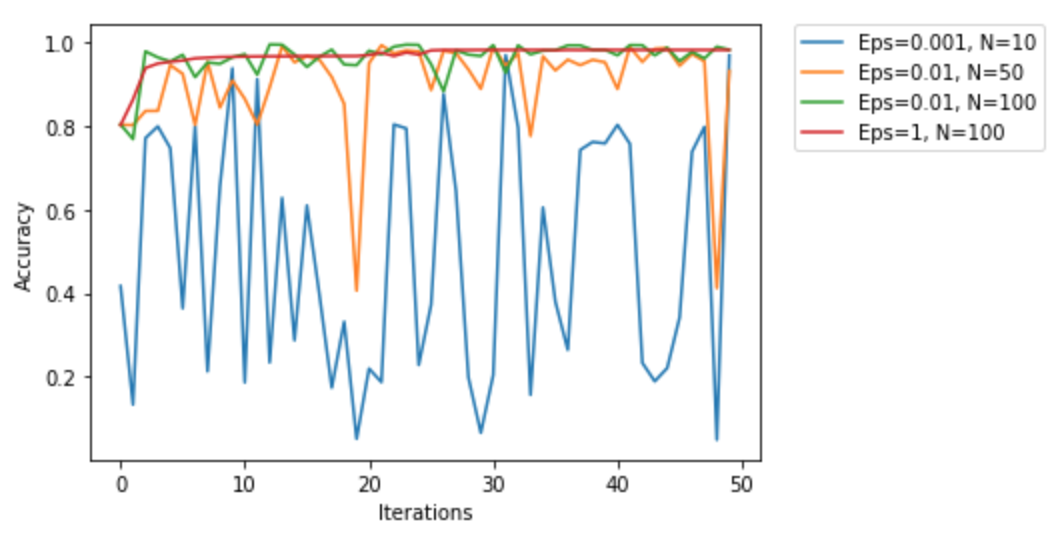}
    \caption{Accuracy vs Iterations for different values of $\epsilon$ and number of parties}
    \label{ResultsTable}
\end{figure}

In the following example, we simulate a larger number of clients. We used the KDDCup99 dataset \cite{bay2000uci} and Algorithm \ref{alg:GoogleclientDP}. 
In Figure \ref{ResultsTable}, one can also see the result of varying the number of clients and $\epsilon$. As expected, the accuracy of the federated model increases with $\epsilon$ since less noise is added. Also, accuracy increases as the number of clients increases. This behavior is expected since more clients contributing to the federated model should increase the accuracy of the federated model. We can also see that convergence happens at a faster rate when the number of parties increases. 

\subsection{Experiment 2: Privacy Constraints }

This experiment illustrates a use case where each client can have different amounts of data and privacy requirements. Suppose Clients 1 and 2 each have 150 data points and $\epsilon = 1.0$. However, Client 3 has 250 data points but also a more stringent privacy requirement, $\epsilon = 0.1$. Clients 1 and 2 may want to decide whether they should only collaborate among themselves, include Client 3, or participate in federated learning at all. 
\begin{itemize}
    \item Scenario 1. No clients participate in federated learning so their accuracy is their own model's accuracy.
    \item Scenario 2. Clients 1 and 2 participate in the federated learning but do not include Client 3 because Client 3 has $\epsilon = 0.1$ while Clients 1 and 2 have $\epsilon = 1.0$.
    \item Scenario 3. All three clients participate in the federated learning and use Client 3's privacy requirement, $\epsilon = 0.1$.
    \item Scenario 4. All three clients participate in federated learning, but Clients 1 and 2 use their privacy requirement $\epsilon = 1.0$, and Client 3 uses its privacy requirement, $\epsilon = 0.1$
\end{itemize}

\subsection{Experiment 3: Decentralized (Serverless) Federated Learning}
\begin{figure}[h!]
    \centering
    \includegraphics[width=0.75\linewidth]{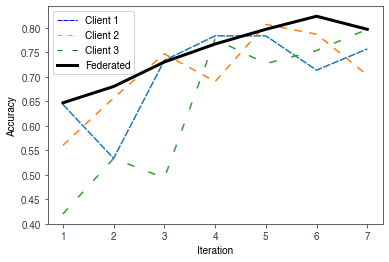}
    \caption{Simulation with Algorithm \ref{alg:cumulativeDP} but modified so that no server is required.}
    \label{Serverless}
\end{figure}

The simulation can also be easily modified beyond the configuration parameters. In the following example, we sub-classed the Agent class to create a new kind of Agent that is able to perform federated machine learning without a server. In this case, clients send their masked weights directly to the other clients. Once a client has received the weights from all the other clients, it is able to average them to create a federated model. Figure \ref{Serverless} shows the results of three clients each training according to Algorithm \ref{alg:cumulativeDP}, with thirty data points per iteration for seven iterations.  In this example, clients use the security protocol but no differential privacy. This behavior is achieved by setting the USE\_DP\_PRIVACY flag to False and the USE\_SECURITY flag to True. As one can see from Figure \ref{Serverless}, all clients clearly benefit from participating in the federated learning in this situation.

\subsection{Experiment 4: Real-World Latency Simulation }
For this experiment, we simulated the situation where the clients are in different geographic locations. We configured three clients as if they were in the Boston, Singapore, NYC, with the server located closest to the NYC client. As such, we designated the following communication latencies: (i) Boston-Server: 0.3 seconds, (ii) Singapore-Server: 2.0 seconds, (iii) NYC-Server: 0.1 seconds.

One can also set client-client latencies. In this example, those only come into effect for the offline portion of the simulation since there is no client-client communication during the online portion of the simulation. 
\begin{center}
\begin{table}
\begin{tabular}{|c|c|c|c|}
\hline
\multirow{2}*{Iteration} & \multicolumn{3}{c|}{\parbox{4.5cm}{Simulated Time to Receive Federated Weights (seconds)}} \\
\cline{2-4}
            & Boston & Singapore & NYC  \\ 
\cline{2-4} 
Iteration 1 & 4.310 & 6.010 & 4.110  \\
Iteration 2 & 4.306 & 6.006 & 4.106   \\
Iteration 3 & 0.909 & Dropped Out &  0.709  \\
\hline
\end{tabular}
\caption{Simulated time to receive federated weights by iteration in an example where Singapore, the farthest client, drops out after the second iteration.}\label{SimulatedTime}
\end{table}
\end{center}

Table \ref{SimulatedTime} shows the simulated time, which measures the time from the start of each iteration to the time when clients would receive the federated weights from the server. The simulated time is available for any step in the protocol since it is included in every message between clients. Upon receiving a message, an agent computes how long its logic has taken. The agent then adds the time that its logic as taken as well as the simulated communication time for whomever the client is sending the message to. This produces a new simulated time for the receiving agent. In Table \ref{SimulatedTime} one can also see that the simulated times for Boston and NYC are significantly lower on the third iteration. This is because clients were permitted to drop out since the CLIENT\_DROPOUT flag was set to True. The Singapore-Server latency is the highest, which means that once Singapore drops out at the end of the second simulation the other clients receive the federated weights faster since the server does not need to wait on the Singaporean client's weights. Similarly, the time taken by each client to compute its weights on each iteration is also displayed by the simulator.

\section{Configuration Parameters}
\label{configParams}
The configuration parameters are described in Table \ref{table:config}
\begin{table}
\small
\begin{tabular}{| c | p{0.5\linewidth} |}
\hline
 USE\_SECURITY & If True, the clients perform a Diffie-Hellman key exchange in the offline portion of the simulation to establish common keys for encryption. Has no effect on federated accuracy.\\ 
 \hline
 USE\_DP\_PRIVACY & If True, clients will add differentially private noise with parameters specified in the config file.  \\  
 \hline
 SUBTRACT\_DP\_NOISE & If True, clients will subtract the noise they added to the federated model upon receiving it from the server. When False, clients use the federated model computed by the server.  \\
 \hline
 CLIENT\_DROPOUT & If True, a client drops out of simulation when each weight in the federated model is within config.tolerance of the client's weight. The simulation continues without that client. \\
 \hline
 SIMULATE\_LATENCIES & If True, the system simulates how long it would take for each step in the protocol to complete using the user-defined communication latencies in config.LATENCY\_DICT. If False, this information is not displayed. \\
 \hline
USING\_CUMULATIVE & If using our data partitioning module, this flag is useful for experimenting between dataset options. If False, the dataset for each iteration includes only the new data available that iteration. This makes sense for Algorithm \ref{alg:GoogleclientDP}. If True, the size of the dataset grows each iteration by the amount of new data. This configuration makes more sense for Algorithm \ref{alg:cumulativeDP}, where the weights of the $i$th iteration are not used in producing the weights of the $i+1$th  iteration.

\\
 \hline
\end{tabular}
\caption{Configuration Parameters}

\label{table:config}
\end{table}

\end{document}